\documentclass{article} 

\usepackage[utf8]{inputenc}

\usepackage{url}
\usepackage{multirow}
\usepackage{lipsum}
\usepackage{amsthm}
\usepackage{mathtools}
\usepackage{xfrac}
\usepackage[export]{adjustbox}
\usepackage{longtable}
\usepackage{booktabs}
\usepackage{colortbl}
\usepackage[dvipsnames]{xcolor}
\usepackage{graphicx}
\usepackage{subcaption}
\usepackage{changelog}
\usepackage{chronosys}
\usepackage{verbatim}
\usepackage{placeins}
\usepackage{tikz}
\usepackage{pgfplots}

\usepackage{svg}
\usepackage{hyperref}
\hypersetup{
    colorlinks=true,
    urlcolor=[rgb]{0.188, 0.498, 0.886},
    linkcolor=[rgb]{0.188, 0.498, 0.886},    
    filecolor=magenta,      
    citecolor=[rgb]{0.188, 0.498, 0.886},
}
\usepackage{fancyvrb}
\usepackage{fixltx2e}
\usepackage{bera}
\usepackage{pdfpages}
\usepackage{pgfgantt}
\usepackage{xcolor}

\definecolor{barblue}{RGB}{153,204,254}
\definecolor{groupblue}{RGB}{51,102,254}
\definecolor{linkred}{RGB}{165,0,33} 
\definecolor{accelblue}{HTML}{4d4cf5}

\usepackage{pgf-umlsd}
\usepackage[draft=true]{minted}

\catcode`\@=11
\def\chron@selectmonth#1{\ifcase#1\or January\or February\or March\or April\or%
 May\or June\or July\or August\or September\or October\or November\or December\fi}

\newcommand\textbluebf[1]{\textbf{\textcolor{accelblue}{#1}}}

\usepackage{algorithm} 
\usepackage{program} 
\usepackage{algorithmic}
\usepackage{listings}
\usepackage{geometry}
\usepackage{authblk}

\usepackage{caption,setspace}
\DeclareCaptionFormat{listing}{\rule{\dimexpr0.9\columnwidth+17pt\relax}{0.4pt}\par\vskip1pt#1#2#3}
\newcounter{code}



\DeclareMathVersion{sans}
\SetSymbolFont{operators}{sans}{OT1}{cmbr}{m}{n}
\SetSymbolFont{letters}  {sans}{OML}{cmbrm}{m}{it}
\SetSymbolFont{symbols}  {sans}{OMS}{cmbrs}{m}{n}

\lstnewenvironment{sflisting}[1][]
  {\lstset{#1}\mathversion{sans}}{}

\usepackage[normalem]{ulem}

\usetikzlibrary{%
  arrows,%
  shapes.misc,
  shapes.arrows,%
  chains,%
  matrix,%
  positioning,
  scopes,%
  decorations.pathmorphing,
  shadows%
}

\definecolor{grayalias}{HTML}{3F4444}

\definecolor{bluealias}{HTML}{307FE2}

\title{\textbf{ROS 2} on a \textbluebf{{\color{accelblue}Chip}}, Achieving Brain-Like Speeds and Efficiency in Robotic Networking}

\author{
    Víctor Mayoral-Vilches, Juan Manuel Reina-Muñoz,
    
    Martiño Crespo Álvarez and David Mayoral-Vilches \\
   \small Acceleration Robotics \\
   \texttt{\small contact@accelerationrobotics.com} \\
}

\newcommand{\roschip}{\emph{ROS 2 on a Chip} }

\begin{document}

\date{} 
\maketitle

\vspace{-1em}
\begin{abstract}


The Robot Operating System (ROS) pubsub model played a pivotal role in developing sophisticated robotic applications. However, the complexities and real-time demands of modern robotics necessitate more efficient communication solutions that are deterministic and isochronous. This article introduces a groundbreaking approach: embedding ROS 2 message-passing infrastructure directly onto a specialized hardware chip, significantly enhancing speed and efficiency in robotic communications. Our FPGA prototypes of the chip design can send or receive packages in less than 2.5 microseconds, accelerating networking communications by more than 62$\times$ on average and improving energy consumption by more than 500$\times$ when compared to traditional ROS 2 software implementations on modern CPUs.  Additionally, it dramatically reduces maximum latency in ROS 2 networking communication by more than 30,000$\times$. In situations of peak latency, our design guarantees an isochronous response within 11 microseconds, a stark improvement over the potential hundreds of milliseconds reported by modern CPU systems under similar conditions.


\end{abstract}


%
%
%

\section{Introduction}
\label{sec:intro}

As robots permeate various sectors - from healthcare to manufacturing - the need for swift and energy-efficient communications becomes paramount. Traditional computational methods, including CPUs and GPUs, are increasingly inadequate for real-time robotics networking communications due to their unbounded intra- and inter-network communication constraints. Robot-specific circuitry can lead to thousands-fold faster communications and better energy-efficiency, growing closer to the realm of human brain-like speeds and efficiency.

In the fascinating realm of neuroscience, a single neuron firing an action potential in the human brain lasts approximately 2 to 5 milliseconds, consuming an incredibly minute amount of energy — roughly 0.03 to 0.3 microjoules. These latencies and energy-efficiency set a benchmark that current message-passing infrastructures in robotics strive to match. The most popular one is the Robot Operating System (ROS)\cite{quigley2009ros}. The most modern implementations of ROS (ROS 2), which build upon the Data Distribution Service (DDS) \cite{pardo2003omg}  operate within similar latency margins on average, however these systems often fall short in maintaining isochronous communications and adhering to comparable energy profiles. The prevalent approach in the industry to meet real-time demands involves an intensive, empirical, and piecemeal tuning of systems. This CPU '\emph{whack-a-mole}' strategy, while common in widely-used robotic frameworks such as ROS 2, proves unsustainable and challenging to scale in industry, especially given the real-time nature of robotic systems.

In this paper, we introduce a pioneering approach to addressing the complexities and real-time demands of modern robotics communication by embedding ROS 2 message-passing infrastructure  functionalities onto specialized hardware chips. This method significantly enhances the speed and efficiency of robotic communications, aligning closer to the rapid and energy-efficient communication seen in the human brain. Our chip design prototyped with an FPGA achieve exceptional improvements in networking communications, offering over 62$\times$ faster message transmission and over 500$\times$ better energy efficiency compared to best performing conventional CPU-based ROS 2 implementations. Moreover, our approach dramatically reduces the maximum latency in ROS 2 networking communication by up to more than 30,000$\times$, ensuring isochronous responses within 11 microseconds even in peak latency situations. This advancement marks a significant leap forward in robotic communications, moving towards more efficient, reliable, and scalable robotic systems.


\section{Background}
\label{sec:background}

In the dynamic world of robotics, ROS emerges as a cornerstone for crafting complex robotic systems, yet faces challenges in meeting the real-time and deterministic communication demands of contemporary robotics. This backdrop fosters a shift towards FPGA-based solutions, aiming to enhance compute and communication speed, efficiency and ensure time-sensitive operations. Research contributions, spanning from foundational tools facilitating FPGA integration within ROS ecosystems \cite{yamashina2016crecomp, 9415584, lienen2020reconros, eisoldt2021reconfros}, to endeavors optimizing the networking stack for improved node interactions \cite{sugata2017acceleration}, delineate a multifaceted approach towards surmounting communication bottlenecks \cite{gutierrez2018real}. Further exploration reveals practical applications \cite{ohkawa2016architecture, panadda2021low, 8956928, 9397897, ohkawa2018fpga, 9355892}, where ROS's computational graph benefits from FPGA-based acceleration, underpinning the pivotal role of hardware acceleration. This narrative underscores the necessity for a comprehensive strategy that not only accelerates user-level applications and ROS libraries but also optimizes the interplay among ROS \texttt{Nodes} across various levels of interaction: inter-process, intra-process, intra-network and inter-network. Such a holistic approach promises substantial latency reductions across ROS and ROS 2's computational landscapes, heralding a new era of efficiency in robotic computation and communication. Tables \ref{table:pastros} covers past research efforts around FPGA-based computing in in ROS and ROS 2.

\startchronology[startyear=2013,stopyear=2024,height=1.0ex]

\chronoevent[textwidth=3.5cm]{2015}{\scriptsize (\textbf{ROS}) Proposal of ROS-compliant fpga component for low-power robotic systems \cite{yamashina2015proposal}}

\chronoevent[textwidth=5cm,markdepth=60pt]{09/2016}{\scriptsize  (\textbf{ROS}) cReComp: Automated Design Toolfor ROS-Compliant FPGA Component \cite{yamashina2016crecomp}}

\chronoevent[textwidth=2.5cm,markdepth=-70pt]{10/2016}{\scriptsize  (\textbf{ROS}) Architecture Exploration of Intelligent Robot System using ROS-compliant FPGA Component  \cite{ohkawa2016architecture}}

\chronoevent[textwidth=5.5cm,markdepth=-20pt]{2017}{\scriptsize (\textbf{ROS}) Acceleration of publish/subscribe messaging in ROS-compliant FPGA component \cite{sugata2017acceleration}}

\chronoevent[textwidth=2.5cm,markdepth=-70pt]{05/2019}{\scriptsize (\textbf{ROS}) High Level Synthesis of ROS Protocol Interpretation and Communication Circuit for FPGA \cite{ohkawa2019high}}

\chronoevent[textwidth=4cm]{12/2019}{\scriptsize (\textbf{ROS}) FPGA-ROS: Methodology to Augment the Robot Operating System with FPGA Designs \cite{podlubne2019fpga}}

\chronoevent[textwidth=4cm,markdepth=120pt]{12/2020}{\scriptsize  (\textbf{ROS 2}) Automated Integration of High-Level Synthesis FPGA Modules with ROS2 Systems \cite{9415584}}

\chronoevent[textwidth=4cm,markdepth=70pt]{12/2020}{\scriptsize  (\textbf{ROS 2}) ReconROS: ReconROS: Flexible Hardware Acceleration for ROS2 Applications \cite{lienen2020reconros}}

\chronoevent[textwidth=3cm,markdepth=-20pt]{02/2021}{\scriptsize (\textbf{ROS}) ReconfROS: Running ROS on Reconfigurable SoCs \cite{eisoldt2021reconfros}}


\chronoevent[textwidth=3cm,markdepth=-80pt]{5/2022}{\scriptsize (\textbf{ROS 2}) RobotCore: An Open Architecture for Hardware Acceleration in ROS 2 \cite{mayoral2022robotcore}}

\chronoevent[textwidth=3cm,markdepth=100pt]{5/2024}{\scriptsize (\textbf{ROS 2}) ROS 2 on a Chip (\textbluebf{this contribution})}

\chronoevent[textwidth=3cm,markdepth=20pt]{10/2023}{\scriptsize (\textbf{ROS 2}) fpgadds: An Intra-FPGA Data Distribution Service for ROS2 Robotics Applications \cite{lienen2023fpgadds}}

\stopchronology

\vspace{2em}

\setlength{\tabcolsep}{10pt}
\renewcommand{\arraystretch}{2.0}

\begin{table}[h!]
    \centering
    \scalebox{0.75}{
    \begin{tabular}{ | >{\centering\arraybackslash}m{1.5cm} | >{\centering\arraybackslash}m{14cm} | >{\centering\arraybackslash}m{1cm} | >{\centering\arraybackslash}m{3cm} | } 
        \hline
        \textbf{Target} & \textbf{Article} & \textbf{Year} & \textbf{Group} \\
        \hline
        ROS & Proposal of ROS-compliant FPGA component for low-power robotic systems \cite{yamashina2015proposal}: First proposal of FPGA ROS-compliant component to simplify FPGA integration into robots. Demonstrated using image processing on Xilinx Zynq 7000, achieving 70\% faster results than running on ARM cores. & 2015 & Utsunomiya University  \\
        \hline
        ROS & cReComp: Automated Design Tool for ROS-Compliant FPGA Component \cite{yamashina2016crecomp}: Proposes an automated tool, cReComp, for creating ROS-compliant FPGA components, enhancing productivity with a code generator and componentization framework. & 2016 & Utsunomiya University \\
        \hline
        ROS & Architecture Exploration of Intelligent Robot System using ROS-compliant FPGA Component \cite{ohkawa2016architecture}: Explores a Visual SLAM study case using ROS-compliant FPGA components and the cReComp design tool. & 2016 & Utsunomiya University \\
        \hline
        ROS & Acceleration of publish/subscribe messaging in ROS-compliant FPGA component \cite{sugata2017acceleration}: Reduces communication latency between ROS components by using hardware accelerated networking stack (SiTCP) and improving ROS API communications. & 2017 & Utsunomiya University \\
        \hline
        ROS & High Level Synthesis of ROS Protocol Interpretation and Communication Circuit for FPGA \cite{ohkawa2019high}: Enhances the ROS-compliant component concept with detailed component descriptions, HLS design flow, and feasibility study. & 2019 & Utsunomiya University, Tokai University, Tokyo City University \\
        \hline
        ROS & FPGA-ROS: Methodology to Augment the Robot Operating System with FPGA Designs \cite{podlubne2019fpga}: Introduces a methodology for designing ROS-Compatible FPGA systems, featuring automatic generation of PL-equivalents to ROS messages and full hardware implementation of ROS entities. & 2019 & Technische Universität Dresden \\
        \hline
        ROS 2 & ReconROS: Flexible Hardware Acceleration for ROS2 Applications \cite{lienen2020reconros}: Presents a framework that integrates ROS with ReconOS for hardware acceleration of ROS applications, enabling ROS nodes fully mapped to hardware. & 2020 & Paderborn University, University of Klangenfurt \\
        \hline
        ROS 2 & Automated Integration of High-Level Synthesis FPGA Modules with ROS2 Systems \cite{9415584}: Introduces the concept of ROS2-FPGA Nodes and the FOrEST tool for seamless integration of HLS-generated FPGA logic into ROS 2 systems, targeting PYNQ 2.5 and achieving significant speed-ups. & 2020 & University of Toronto, Shibaura Institute of Technology, Keio University, Tokai University \\
        \hline
        ROS & ReconfROS: Running ROS on Reconfigurable SoCs \cite{eisoldt2021reconfros}: Proposes a method to integrate ROS into adaptive SOCs, focusing on general algorithmic acceleration and easy integration of FPGA-based hardware accelerators into ROS nodes. & 2021 & Osnabrück University \\
        \hline
        ROS 2 & RobotCore: An Open Architecture for Hardware Acceleration in ROS 2 \cite{mayoral2022robotcore}: Introduces RobotCore, a hardware acceleration robotics framework that demonstrates an intra-FPGA ROS 2 node communication queue to enable faster dataflows, and use it in conjunction with FPGA-accelerated nodes to achieve a 24.42\% speedup over a CPU. & 2022 & Acceleration Robotics, Harvard Unversity and Columbia University \\
        \hline
        ROS 2 & fpgadds: An Intra-FPGA Data Distribution Service for ROS2 Robotics Applications \cite{lienen2023fpgadds}: Demonstrates fpgaDDS, a customized streaming communication architecture for low latency and high bandwidth between hardware-mapped ROS nodes, showing a 13x performance improvement over software DDS. & 2023 & Paderborn University \\
        \hline
    \end{tabular}}
    \caption{Past research efforts to bring ROS into FPGAs.}
    \label{table:pastros}    
\end{table}

Given the emerging interest and demonstrated advantages, the exploration of specialized circuitry for accelerating ROS 2's message-passing infrastructure, particularly through the Data Distribution Service (DDS), marks a pivotal advancement towards ultra-efficient robotic communications. This initiative aims to enable robotic systems to exchange information in just a few microseconds, significantly outperforming the latency benchmarks set by current software implementations of ROS 2. The subsequent section introduces our work to bring \emph{"ROS 2 on a Chip"} and will elaborate on the integration of ROS 2 functionalities into specialized hardware, focusing on the potential of FPGA-based accelerations to significantly enhance real-time capabilities and computational efficiency in robotic systems.

\section{ROS 2 on a Chip}
\label{sec:rosonachip}

The architecture of robot communication stands as a critical foundation for advanced computation and operational functionality. At the heart of this architecture lies the computational graph, a model pioneered by the Robot Operating System (ROS), which has become a standard for developing intricate robotic systems. This graph structures the flow of data from an array of sensors, weaving through a network of computational nodes that process, analyze, and act upon this information in a coordinated fashion. As such, the dataflow within and across robotic systems encapsulates the essence of robot cognition, making intra-robot communications indispensable to the entirety of robotic computations. Recognizing the significance of this communication infrastructure, our investigation delves into the advancement of \roschip — an initiative aimed at embedding ROS 2 functionalities into specialized hardware. 




\FloatBarrier

\begin{figure}[h!]
    \vspace{.5em}
    \centering
    \includegraphics[width=1.0\textwidth]{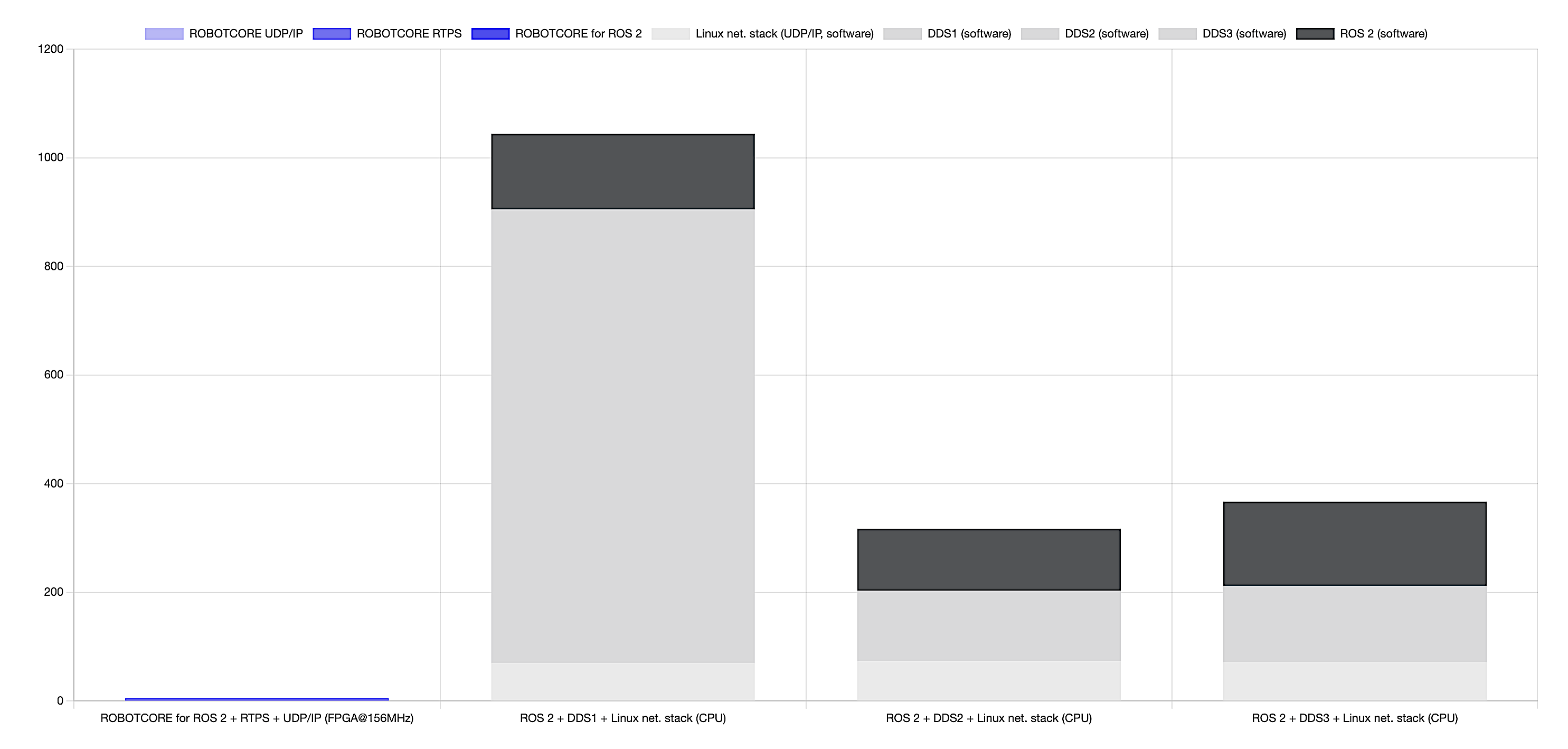}

    \setlength{\tabcolsep}{20pt}
    \renewcommand{\arraystretch}{1.5}
        \scalebox{0.6}{
        \begin{tabular}{  >{\centering\arraybackslash}m{5cm} |>{\centering\arraybackslash}m{4cm}|>{\centering\arraybackslash}m{4cm}|>{\centering\arraybackslash}m{4cm}|>{\centering\arraybackslash}m{4cm} } 
            
              & \color{black}\textbluebf{ROBOTCORE for ROS 2 + RTPS + UDP/IP (}\texttt{FPGA@156MHz}\textbluebf{)} & \color{black}\textbf{ROS 2 + DDS1 + Linux net. stack (CPU)} & \color{black}\textbf{ROS 2 + DDS2 + Linux net. stack (CPU)} & \color{black}\textbf{ROS 2 + DDS3 + Linux net. stack (CPU)} \\
            \hline
            
            ROBOTCORE UDP/IP (\texttt{hardware, FPGA@156MHz}) & \textbluebf{0.7} & & &  \\ 
            \rowcolor{black!5} ROBOTCORE RTPS (\texttt{hardware, FPGA@156MHz}) & \textbluebf{2.3} & & &  \\ 
            ROBOTCORE for ROS 2 (\texttt{hardware, FPGA@156MHz}) & \textbluebf{2} & & &  \\
            \rowcolor{black!5} Linux UDP/IP net. stack (\texttt{software, CPU}) &  & 70 & 73 & 71 \\  
            DDS1 (\texttt{software, CPU}) &  & 835 &  &   \\
            \rowcolor{black!5} DDS2 (\texttt{software, CPU}) &  &  & 127 &   \\
            DDS3 (\texttt{software, CPU}) &  &  &  & 141  \\
            \rowcolor{black!5} ROS 2 (\texttt{software, CPU}) &  & 139 & 114 & 155  \\
    
            \hline
            \textbf{Total in us} (\texttt{speedup}) & \textbluebf{5} (\texttt{1$\times$}) & \textbf{1044} (\texttt{208$\times$}) & \textbf{314} (\texttt{62$\times$}) & \textbf{369} (\texttt{73$\times$})\\            
        \end{tabular}}
        
        \caption{Mean Round-Trip Network Latency in microseconds (us) breakdown across various combination of hardware and software implementations. Round-Trip Time (RTT) mean latencies measured after 1M samples and while sending small ROS messages in a ping-pong format. ROBOTCORE for ROS 2, ROBOTCORE RTPS and ROBOTCORE UDP/IP running in an FPGA@156MHz. Software implementations running on an AMD Ryzen 5 PRO 4650G.}
        \label{fig:comparison}        

\end{figure}  

We implement the complete ROS 2 robotics communication framework stack split into three hardware designs including all lower abstraction layers in hardware to deliver unprecedented speed and efficiency in ROS 2 interactions over the network\footnote{Note that in this work we focus on the intra-network ROS 2 interactions.}: ROBOTCORE® for ROS 2, ROBOTCORE® RTPS and ROBOTCORE® UDP/IP. Our chip design which includes all three is prototyped in an FPGA and significantly enhances the networking architecture of robotic systems and tackles one the widely criticized issues of ROS 2: its latency overhead over the DDS communication middleware. It does so by building a hardware implementation of the core ROS 2 abstraction layers (RCL, RMW) and by establishing direct hardware datapaths with the underlying DDS middleware implementation (\ref{fig:ros2_architecture}). We then hardware-implement the Real-Time Publish Subscribe protocol (RTPS) \cite{rtps25} with some minimal DDS capabilities for interoperability (also in hardware) (\ref{fig:rtps_dds_architecture}). This allows to remove the latency overhead of ROS 2 over DDS and to deliver unmatched speed and absolute determinism. Finally, we connect the optimized datapaths of our minimal hardware DDS implementation with an ultra low-latency hardware UDP/IP stack that sits on top of a 10G PHY (\ref{fig:udpip_architecture}). 


Results comparing our ROS 2 circuitry with software implementations are depicted in Figure \ref{fig:comparison}, with all software benchmarks conducted on a modern AMD Ryzen 5 PRO 4650G CPU equipped with 64 GB of RAM. Our comprehensive evaluation showcases the exceptional performance of the ROS 2 hardware architecture, significantly surpassing traditional software-based approaches. Notably, our FPGA-based designs—ROBOTCORE® for ROS 2, ROBOTCORE® RTPS, and ROBOTCORE® UDP/IP—demonstrate groundbreaking efficiency, achieving mean latency figures as low as \texttt{0.7} microseconds for UDP/IP, \texttt{2.3} microseconds for RTPS, and an aggregate of \texttt{2} microseconds for ROS 2 functionalities, all at a clock rate of 156 MHz.

In stark contrast, software implementations operating on the CPU exhibit substantially higher latencies, with Linux UDP/IP networking stack alone contributing up to a mean of \texttt{73} microseconds. When considering DDS implementations, mean latency further escalates, with \texttt{DDS1} reaching \texttt{835} microseconds, \texttt{DDS2} at \texttt{127} microseconds, and \texttt{DDS3} at 141 microseconds. Even the aggregated ROS 2 software mean latency ranges from \texttt{114} to \texttt{155} microseconds, dependent on the DDS version used. Consequently, the total latency for our FPGA-based solution stands at an impressive \texttt{5} microseconds of mean latency, marking a speedup factor ranging from \texttt{62$\times$} to \texttt{208$\times$} when compared to various software configurations.

It's crucial to note that all comparative experiments were conducted under optimal conditions for the CPU-based setups: with an idle CPU processor and an idle network. These conditions artificially inflate the performance of the CPU implementations, which, unlike our FPGA-based solution, are not designed for hard real-time or isochronous operations (further covered in subsection \ref{sec:isochronous}). As such, the demonstrated latency figures for software configurations, while already significantly higher than our hardware solution, represent a best-case scenario\footnote{In real-world robotic applications, CPU processors and networks are rarely idle and often experience varying levels of saturation. As demonstrated by previous work at \cite{gutierrez2018real, gutierrez2018towards}, the latency and performance inconsistencies of software-based ROS 2 communications would become more pronounced.}. This remarkable achievement not only underscores the superior performance of hardware acceleration in ROS 2 communications but also highlights the transformative potential of integrating ROS 2 functionalities directly into specialized hardware for robotics applications, setting a new benchmark for real-time robotic communication and computational efficiency.

\subsection{Processor Design}

\begin{figure}[h!]
    \centering
    

    \caption{Resource Utilization Breakdown for the "\roschip" Design on FPGA Platforms: Detailing the allocations for integrated submodules. ROBOTCORE® for ROS 2 facilitates core ROS 2 abstractions; ROBOTCORE® RTPS provides core RTPS message-passing infrastructure and DDS interoperability; and ROBOTCORE® UDP/IP enables high-speed, fixed low-latency networking stack interactions. This composition highlights the comprehensive approach to hardware-accelerated robotic communication}
    
    \label{fig:ros2_chip_resources}
    
    \begin{subfigure}[b]{0.6\textwidth}
        \includegraphics[width=\textwidth]{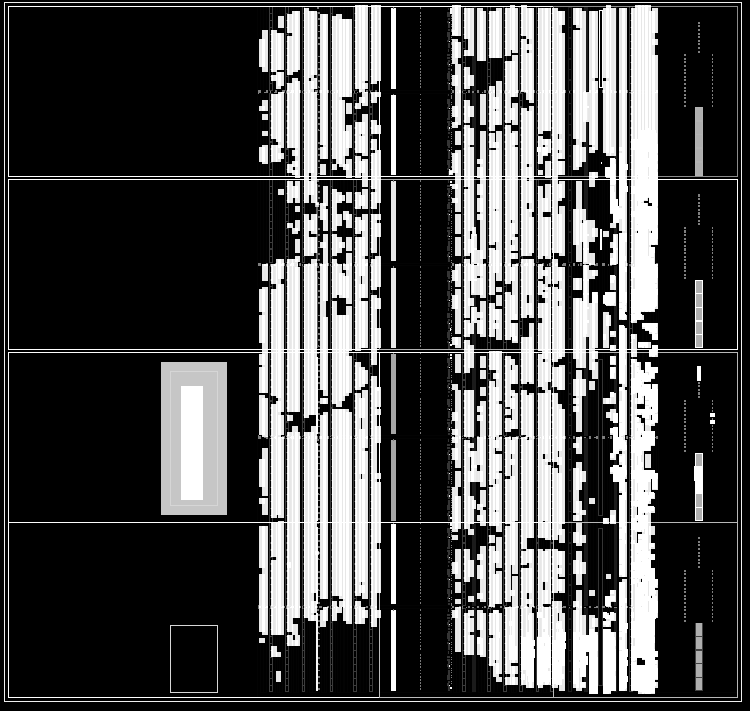}
        \caption{Complete design resource allocation on AMD Zynq™ UltraScale+™ MPSoC.}
        \label{fig:all_resources}
    \end{subfigure}
    \\
    \begin{subfigure}[b]{0.2\textwidth}
        \includegraphics[width=\textwidth]{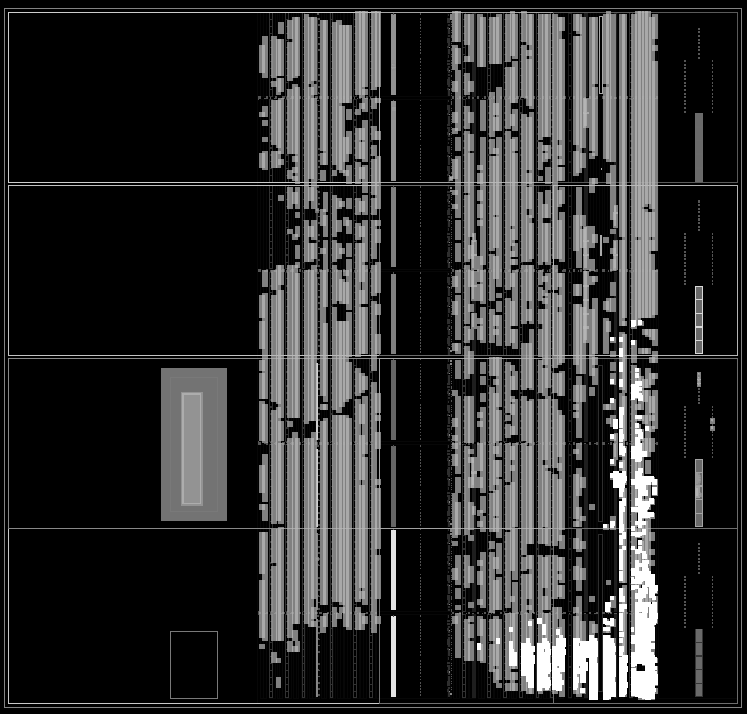}
        \caption{IP and UDP resource allocation on AMD Zynq™ UltraScale+™ MPSoC.}
        \label{fig:ip_udp_resources}
    \end{subfigure}
    ~
    \begin{subfigure}[b]{0.2\textwidth}
        \includegraphics[width=\textwidth]{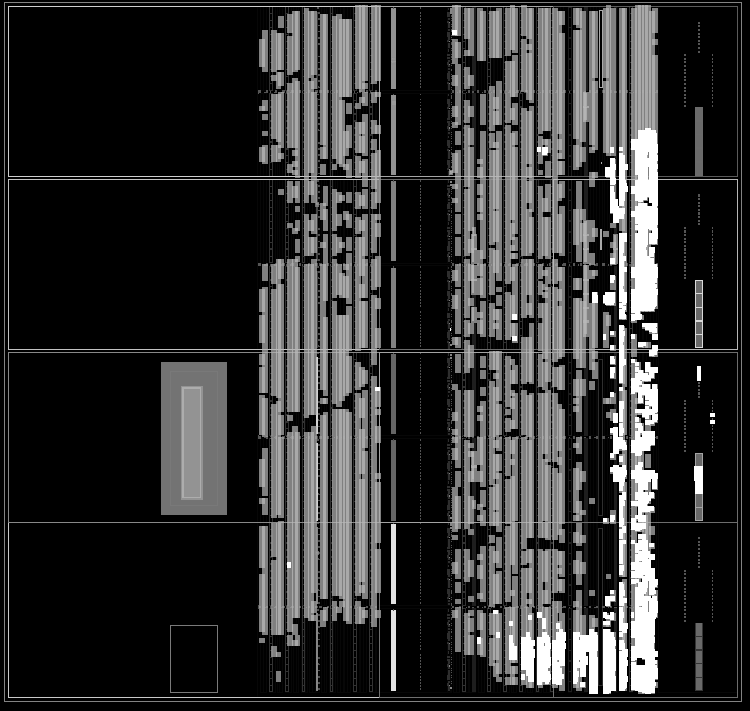}
        \caption{PHY, IP, and UDP resource allocation on AMD Zynq™ UltraScale+™ MPSoC.}
        \label{fig:phy_ip_udp_resources}
    \end{subfigure}
    ~
    \begin{subfigure}[b]{0.2\textwidth}
        \includegraphics[width=\textwidth]{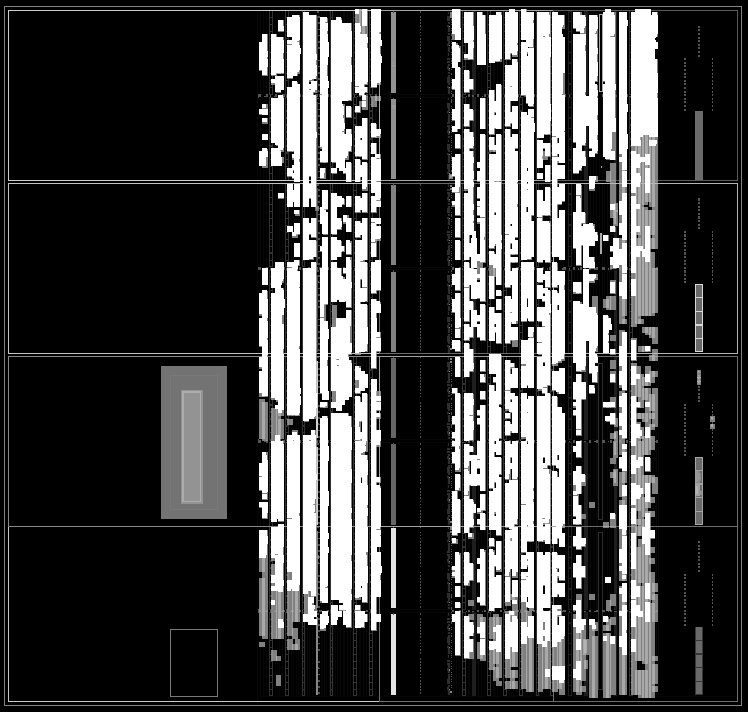}
        \caption{RTPS and ROS 2 resource allocation on AMD Zynq™ UltraScale+™ MPSoC.}
        \label{fig:rtps_ros2_resources}
    \end{subfigure}
    ~
    \begin{subfigure}[b]{0.2\textwidth}
        \includegraphics[width=\textwidth]{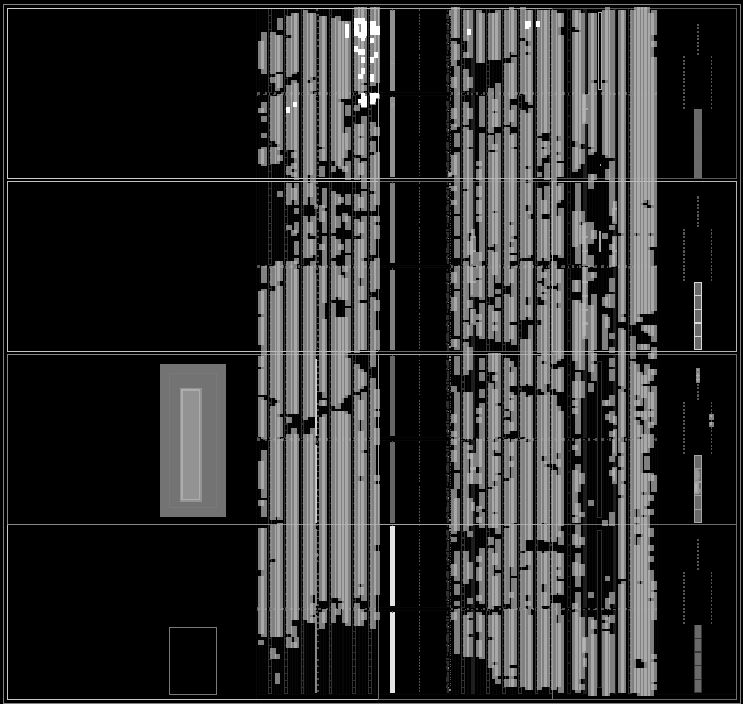}
        \caption{PubSub resource allocation on AMD Zynq™ UltraScale+™ MPSoC.}
        \label{fig:pubsub_resources}
    \end{subfigure}

    \setlength{\tabcolsep}{20pt}
    \renewcommand{\arraystretch}{1.5}
    \scalebox{0.6}{
    \begin{tabular}{  >{\centering\arraybackslash}m{5cm} |>{\centering\arraybackslash}m{3cm}|>{\centering\arraybackslash}m{3cm}|>{\centering\arraybackslash}m{3cm}|>{\centering\arraybackslash}m{3cm} } 
        
        \rowcolor{black}
          & \color{white}\textbf{LUT} (\texttt{\%}) & \color{white}\textbf{FF} (\texttt{\%}) & \color{white}\textbf{DSP} (\texttt{\%}) & \color{white}\textbf{BRAM} (\texttt{\%})\\
        \hline
        
        AMD Zynq™ UltraScale+™ MPSoC EV (XCK26) & 56,440 (\texttt{48.19\%}) & {20,285} (\texttt{8.66\%}) & {0} (\texttt{0\%}) & {26} (\texttt{18.06\%}) \\
        
        
    \end{tabular}} 

    \setlength{\tabcolsep}{20pt}
    \renewcommand{\arraystretch}{1.5}
    \scalebox{0.6}{
    \begin{tabular}{  >{\centering\arraybackslash}m{5cm} |>{\centering\arraybackslash}m{3cm}|>{\centering\arraybackslash}m{3cm}|>{\centering\arraybackslash}m{3cm}|>{\centering\arraybackslash}m{3cm} } 
        
        \rowcolor{black}
          & \color{white}\textbf{ALUT} (\texttt{\%}) & \color{white}\textbf{REG} (\texttt{\%}) & \color{white}\textbf{DSP} (\texttt{\%}) & \color{white}\textbf{M20K} (\texttt{\%})\\
        \hline
        
        
        Intel Agilex® 7 FPGA F-Series, 1400 KLE, 2486A & 187,962 (\texttt{19.29\%}) & 163,115 (\texttt{8.37\%}) & 0 (\texttt{0\%}) & 138 (\texttt{1.94\%}) \\
        
    \end{tabular}}     
\end{figure}

The architecture of \roschip composed by ROBOTCORE® for ROS 2, RTPS and UDP/IP submodules is designed to be highly scalable and flexible, enabling it to seamlessly integrate into a variety of FPGA and FPGA SoC technologies. Figure \ref{fig:ros2_chip_resources}'s detailed portrayal of resource allocation for the \roschip processor design on AMD Zynq™ UltraScale+™ MPSoC and Intel Agilex® 7 FPGA F-Series provides keen insights into the efficient use of FPGA resources. It is noted that the resources required for implementing the chip design consume less than \texttt{50\%} of the total available on the AMD platform and under \texttt{20\%} on the Intel platform. This efficient resource usage underscores the design's scalability and potential for integration into a wide range of applications without nearing resource capacity limits.

\FloatBarrier

A significant observation from the resource allocation breakdown is that the bulk of resources are dedicated to the RTPS and DDS logic, which are pivotal for the communication capabilities of the chip. These components demand the most from the FPGA's resources due to their complexity and the need for high-speed, deterministic communication. In contrast, a smaller portion of the resources is allocated to the parallel UDP/IP networking stack, and even less to the ROS 2 abstractions in hardware, which highlights the lightweight nature of these abstractions and their minimal impact on resource consumption.
The resource sharing for RTPS and DDS logic further benefits the scalability of the chip design, especially as ROS 2 requires minimal additional resources to be implemented in hardware. This means that adding more ROS 2 abstractions such as \texttt{publishers}, \texttt{actions}, or \texttt{services} would require only a marginal increase in hardware space, thereby enhancing the chip's parallelization capabilities and its suitability for more complex ROS 2 applications. This efficient and scalable use of resources not only emphasizes the advanced design of the \roschip but also showcases its potential to impact robotics applications by facilitating more complex computations and interactions within a compact hardware footprint.

\vspace{-1em}

\begin{figure}[h!]
    \centering
    \begin{subfigure}[b]{0.95\textwidth}
        \includegraphics[width=\textwidth]{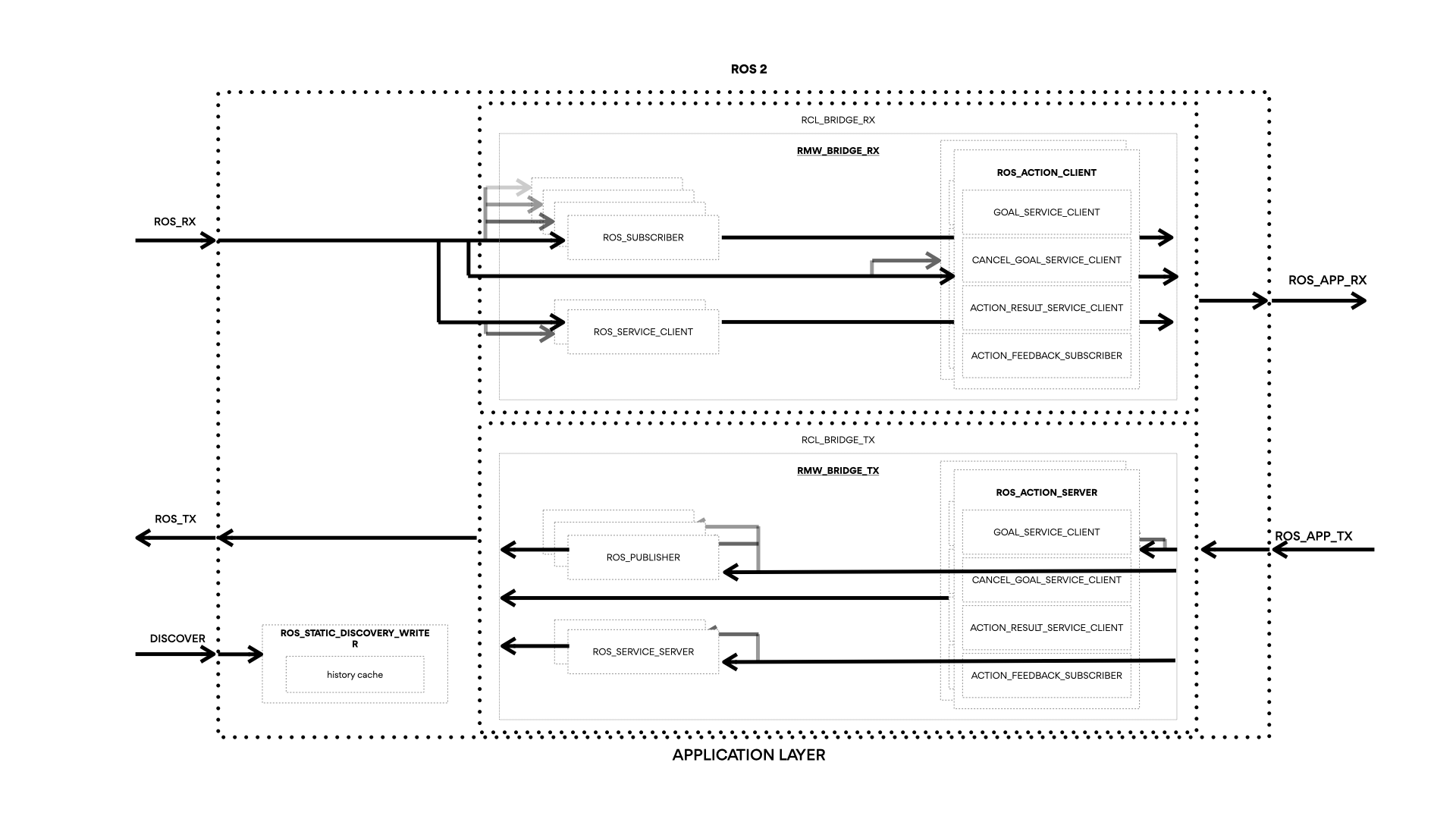}
        \caption{ROS 2 hardware accelerator data flow.}
        \label{fig:ros2_architecture}
    \end{subfigure}
    \hfill 
    \begin{subfigure}[b]{0.47\textwidth}
        \includegraphics[width=\textwidth]{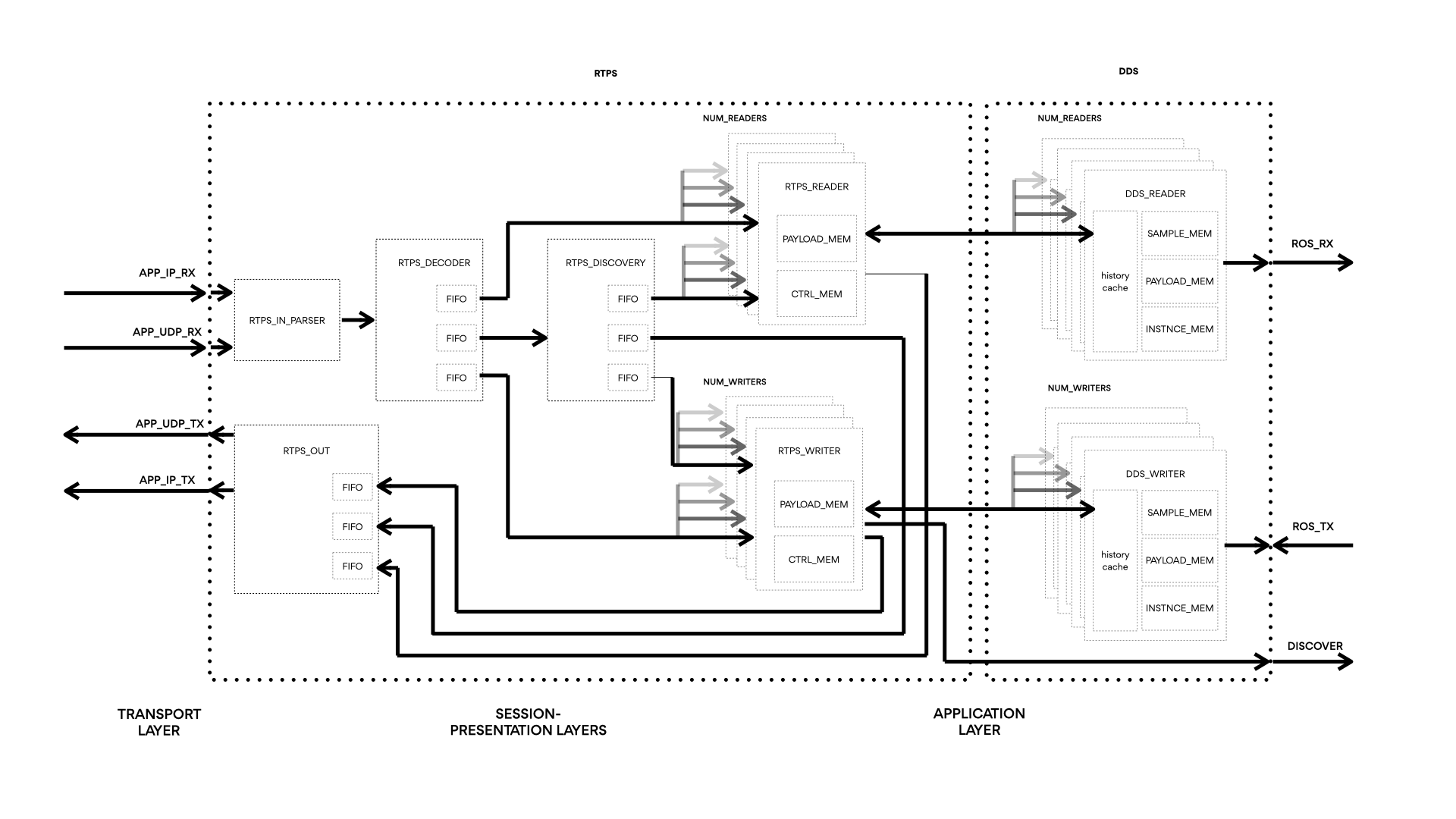}
        \caption{RTPS/DDS hardware accelerator data flow.}
        \label{fig:rtps_dds_architecture}
    \end{subfigure}    
    \begin{subfigure}[b]{0.47\textwidth}
        \includegraphics[width=\textwidth]{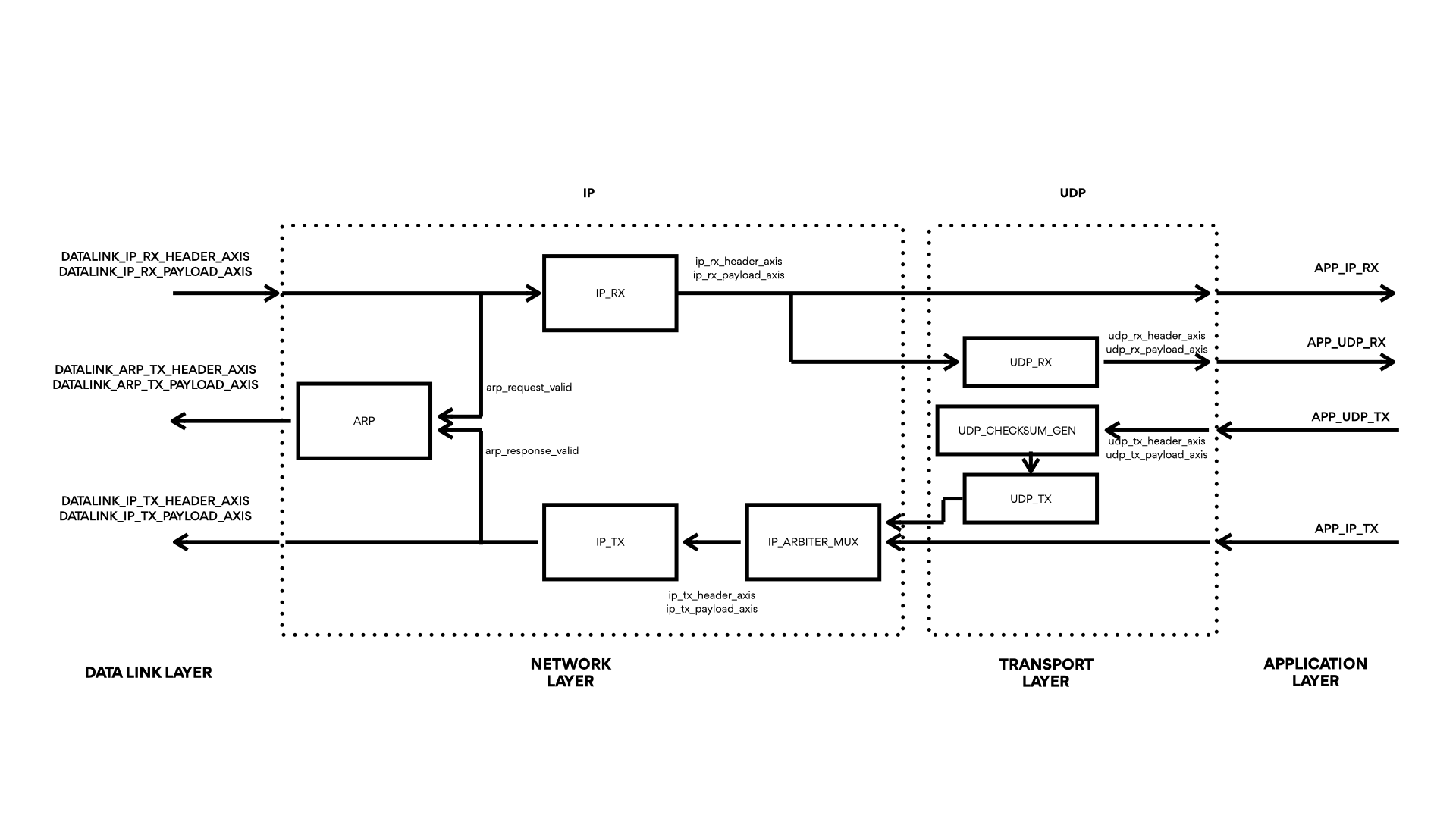}
        \caption{UDP/IP hardware accelerator data flow.}
        \label{fig:udpip_architecture}
    \end{subfigure}
    \hfill 
    \caption{Hardware accelerator data flow diagrams for UDP/IP, RTPS/DDS, and ROS 2.}
    \label{fig:architectures}
\end{figure}

At its core, \roschip implements a streamlined and optimized hardware processor for ROS 2 networking parallelizing computing across various hardware sub-cores, each of which can have multiple instances and is responsible for a specific ROS 2 operation streaming data in parallel. Figure \ref{fig:architectures} depicts the architecture of the various accelerator designs that compose the \roschip circuitry. Subfigure \ref{fig:ros2_architecture} corresponds with ROS 2 abstractions implemented in hardware. Data-paths are fed from the underlying communication middleware (depicted in subfigure \ref{fig:rtps_dds_architecture}) and into the robotic application. Control-paths exist similarly, incoming from the underlying communication middleware and also between blocks of the hardware implementation of this layer. ROS 2-level abstractions implemented in hardware (such as \texttt{PUBLISHERS}, \texttt{SUBSCRIBERS}, \texttt{ACTION\_SERVERS}, etc) can get hardware-instantiated multiple times, which leads to higher throughput at an additional energy budget. Determinism is guaranteed and maximum latency is established by the hardware implementation datapaths, which is limited by the maximum frequency (\texttt{fmax}) of operation, which in the case of the AMD Zynq™ UltraScale+™ MPSoC turned \texttt{156 MHz} and for the Intel Agilex® 7 FPGA F-Series \texttt{161 MHz}. Subfigure \ref{fig:rtps_dds_architecture} depicts an implementation of a hardware DDS communication middleware. Control- and data-paths lie between the relationships across blocks. Readers and writers across RTPS (the interoperability protocol) and DDS (the middleware on top) can be instantiated multiple times for further parallelism while interfacing with upstream robotic applications. Downstream data paths such as \texttt{RTPS\_OUT} are built in hardware, connected to underlying hardware implementation and can also be set redundant and parallel if the application demands it. Isochronous behavior is guaranteed by the hardware datapaths and maximum latency is established by the hardware implementation, which is also limited by the maximum frequency (\texttt{fmax}) of operation. Finally, \ref{fig:udpip_architecture} depicts an architecture diagram of the UDP/IP stack implementated in hardware with dedicated data paths per each networking interaction that run concurrently and without interference. While they run in parallel (interleaving) with all the upstream robotics stack layers implemented also in hardware.

\subsection{Thousands-Fold faster,  deterministic and isochronous}
\label{sec:isochronous}

\begin{figure}[b!]
    \vspace{-1em}
    \centering
    \includegraphics[width=1.0\textwidth]{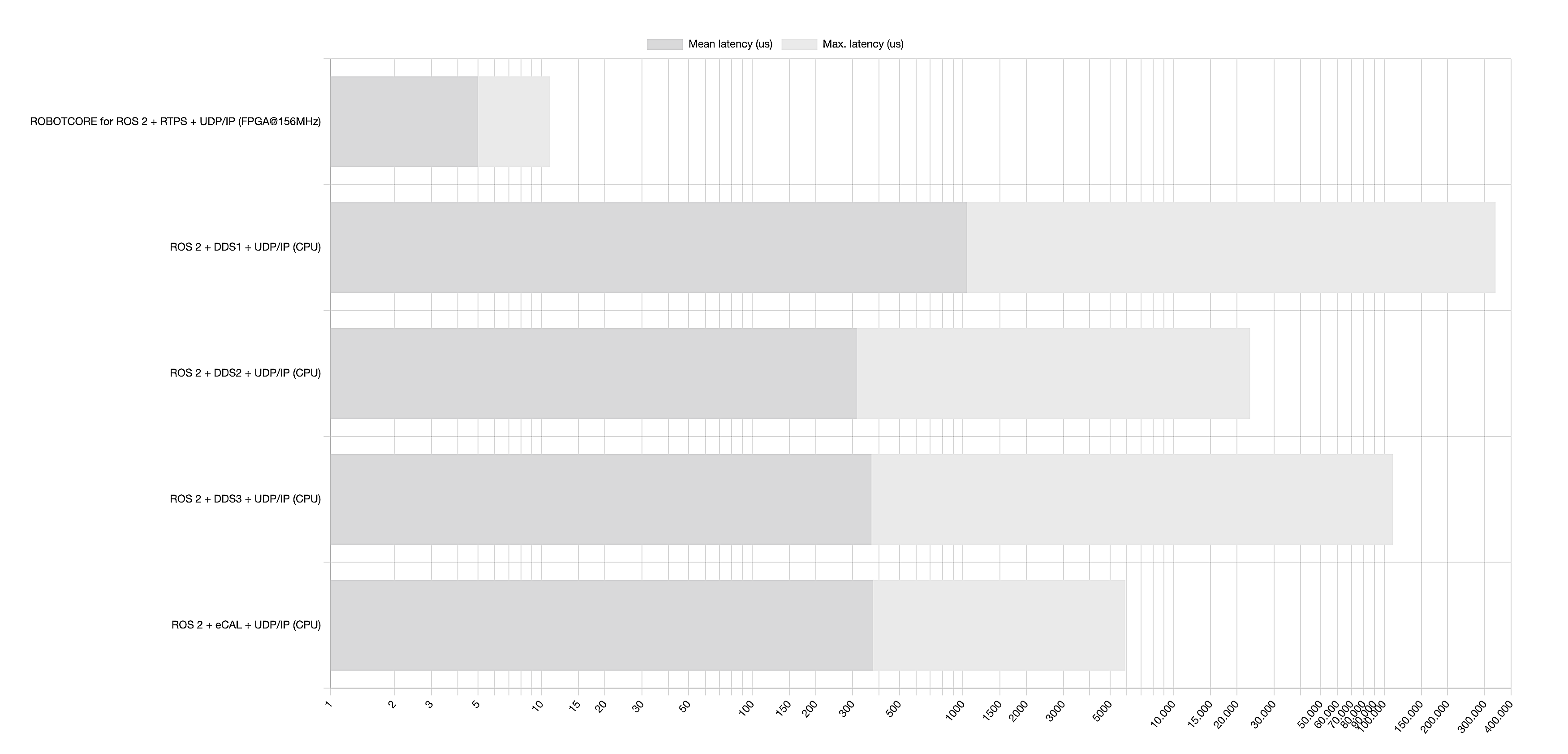}

    \setlength{\tabcolsep}{20pt}
    \renewcommand{\arraystretch}{1.5}
    \scalebox{0.6}{
    \begin{tabular}{  >{\centering\arraybackslash}m{5cm} |>{\centering\arraybackslash}m{2.9cm}|>{\centering\arraybackslash}m{2.9cm}|>{\centering\arraybackslash}m{2.9cm}|>{\centering\arraybackslash}m{2.9cm}|>{\centering\arraybackslash}m{2.9cm} } 
        
          & \color{black}\textbluebf{ROBOTCORE for ROS 2 + RTPS + UDP/IP (}\texttt{FPGA@156MHz}\textbluebf{)} & \color{black}\textbf{ROS 2 + DDS1 + Linux net. stack (CPU)} & \color{black}\textbf{ROS 2 + DDS2 + Linux net. stack (CPU)} & \color{black}\textbf{ROS 2 + DDS3 + Linux net. stack (CPU)} & \color{black}\textbf{ROS 2 + eCAL + Linux net. stack (CPU)} \\
        \hline
        
        \rowcolor{black!5} Mean latency (\texttt{slowdown})& \textbluebf{5} (\texttt{1$\times$}) & {1044} (\texttt{208$\times$}) & {314} (\texttt{62$\times$}) & {369} (\texttt{73$\times$}) & {376} (\texttt{75$\times$}) \\
        
        Max. latency (\texttt{slowdown})& \textbluebf{11} (\texttt{1$\times$}) & \textbf{336750} (\texttt{30613$\times$}) & \textbf{22769} (\texttt{2069$\times$}) & \textbf{109776} (\texttt{9979$\times$}) & \textbf{5532} (\texttt{502$\times$}) \\
        
    \end{tabular}}
    \caption{Mean and Maximum Round-Trip Network Latency in microseconds (us) breakdown across various combination of hardware and software implementations. Logarithmic scale. Figure depicts how a hardware implementation delivers an isochronous and faster interaction.}
    \label{fig:faster_isochronous}
\end{figure} 

The quantitative analysis presented in our study emphasizes not only the remarkable speedup achieved by our \roschip FPGA prototype implementation in terms of mean and maximum latency but also brings to light its isochronous capabilities, which are essential for ensuring time-deterministic communications in robotics. The data, as depicted in Figure \ref{fig:faster_isochronous}, offers a stark comparison between the hardware-accelerated and software-based solutions, highlighting the intrinsic advantages of our FPGA-prototype approach in achieving isochronous behavior.

In detail, our chip implementation prototyped in an FPGA maintains a mean latency of \texttt{5} microseconds and a remarkably low maximum latency of \texttt{11} microseconds, demonstrating the system's capability to deliver consistent performance under varying conditions. This isochronous nature of communication is crucial for real-time robotic applications, where even minor deviations in message timing can lead to significant operational discrepancies. In contrast, software implementations on a CPU, across various DDS configurations and \texttt{eCAL}, exhibit significantly higher and more variable latencies, with maximum latency figures soaring to as high as \texttt{336,750} microseconds for \texttt{DDS1}. This variability undermines the predictability and reliability of software-based communications, rendering them unsuitable for scenarios requiring stringent real-time guarantees. Furthermore, the comparison reveals that our FPGA implementation achieves up to a \texttt{30,613$\times$} improvement in maximum latency over the software configurations, underscoring the substantial enhancement in determinism. This deterministic performance, characterized by minimal latency variations, is absent in the software implementations, where latency can fluctuate dramatically depending on system and network load, further exacerbating the challenge of achieving isochronous communications.

In essence, the FPGA-based "ROS 2 on a Chip" design not only accelerates communication speeds but also solidifies the foundation for isochronous and deterministic robotic communications. This advancement paves the way for more sophisticated and reliable robotic systems, capable of performing synchronized tasks with unparalleled precision and safety. Through the adoption of this hardware-accelerated approach, we address and overcome a critical bottleneck in robotic communications, marking a significant leap towards realizing the full potential of real-time and autonomous robotic systems.

\subsection{100-40000x better than other IP cores}


\begin{figure}[h!]
    \vspace{.5em}
    \centering
    \includegraphics[width=1.0\textwidth]{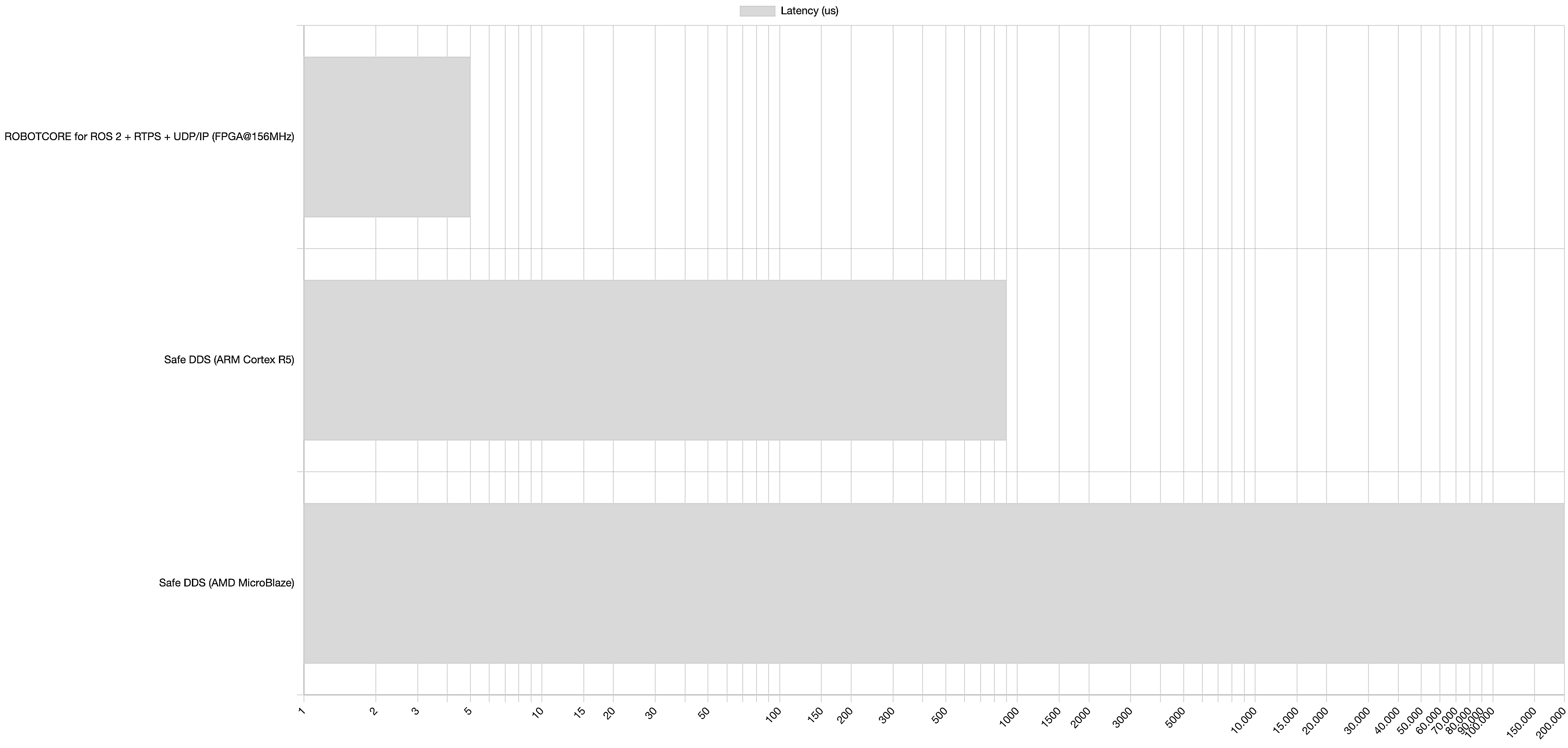}

    \setlength{\tabcolsep}{20pt}
    \renewcommand{\arraystretch}{1.5}
    \scalebox{0.6}{
    \begin{tabular}{  >{\centering\arraybackslash}m{5cm} |>{\centering\arraybackslash}m{4cm}|>{\centering\arraybackslash}m{4cm}|>{\centering\arraybackslash}m{4cm} } 
        
          & \color{black}\textbluebf{ROBOTCORE for ROS 2 + RTPS + UDP/IP (}\texttt{FPGA@156MHz}\textbluebf{)} & \color{black}\textbf{Safe DDS (ARM Cortex-R5)} & \color{black}\textbf{Safe DDS (AMD MicroBlaze)} \\
        \hline
        
        \rowcolor{black!5} Mean Round-Trip Network Latency (\texttt{speedup}) & \textbluebf{5} (\texttt{1$\times$}) & {900} (\texttt{180$\times$}) & {200,000} (\texttt{40,000$\times$}) \\
                
    \end{tabular}}
    \caption{Mean Round-Trip Network Latency in microseconds (us) across various combination of hard and soft-processors providing ROS 2 interoperability capabilities. Logarithmic scale.}
    \label{fig:other_chips}
\end{figure} 

\noindent Our \roschip design, when evaluated against other IP core solutions offering DDS and RTPS functionalities on different processing architectures, showcases distinct performance metrics, particularly in terms of mean round-trip network latency. As presented in Figure \ref{fig:other_chips}, our circuitry implemented on an FPGA operating at a frequency of \texttt{156MHz}, records a mean round-trip network latency of \texttt{5} microseconds. This result is juxtaposed with \texttt{900} microseconds observed for Safe DDS running on an ARM Cortex-R5 and \texttt{200,000} microseconds on an AMD MicroBlaze, indicating latency reductions by factors of \texttt{180} and \texttt{40,000}, respectively\cite{SafeDDS2023}. Such disparities in performance are critical in the context of robotic communications, where latency impacts the system's responsiveness and operational efficiency. Our \roschip FPGA-prototype provides a stark contrast in network latency compared to implementations on both hard-core (ARM Cortex-R5) and soft-core (AMD MicroBlaze) processors. This comparative analysis highlights the significant differences in how each architecture handles ROS 2 interoperability requirements, with our FPGA-based prototype demonstrating the lowest latency figures.

Furthermore, the comparison sheds light on the inherent limitations and performance bottlenecks associated with soft-core and hard-core processor-based deployments for DDS and RTPS functionalities within ROS 2 frameworks. By detailing these quantitative results, the study underscores the potential advantages of a custom compute architecture in reducing network latency for robotic communication systems, thereby enhancing real-time data exchange and processing capabilities essential for autonomous robotic operations.

\subsection{500$\times$ more energy-efficient}

\begin{figure}[h!]
    \vspace{.5em}
    \centering
    \includegraphics[width=1.0\textwidth]{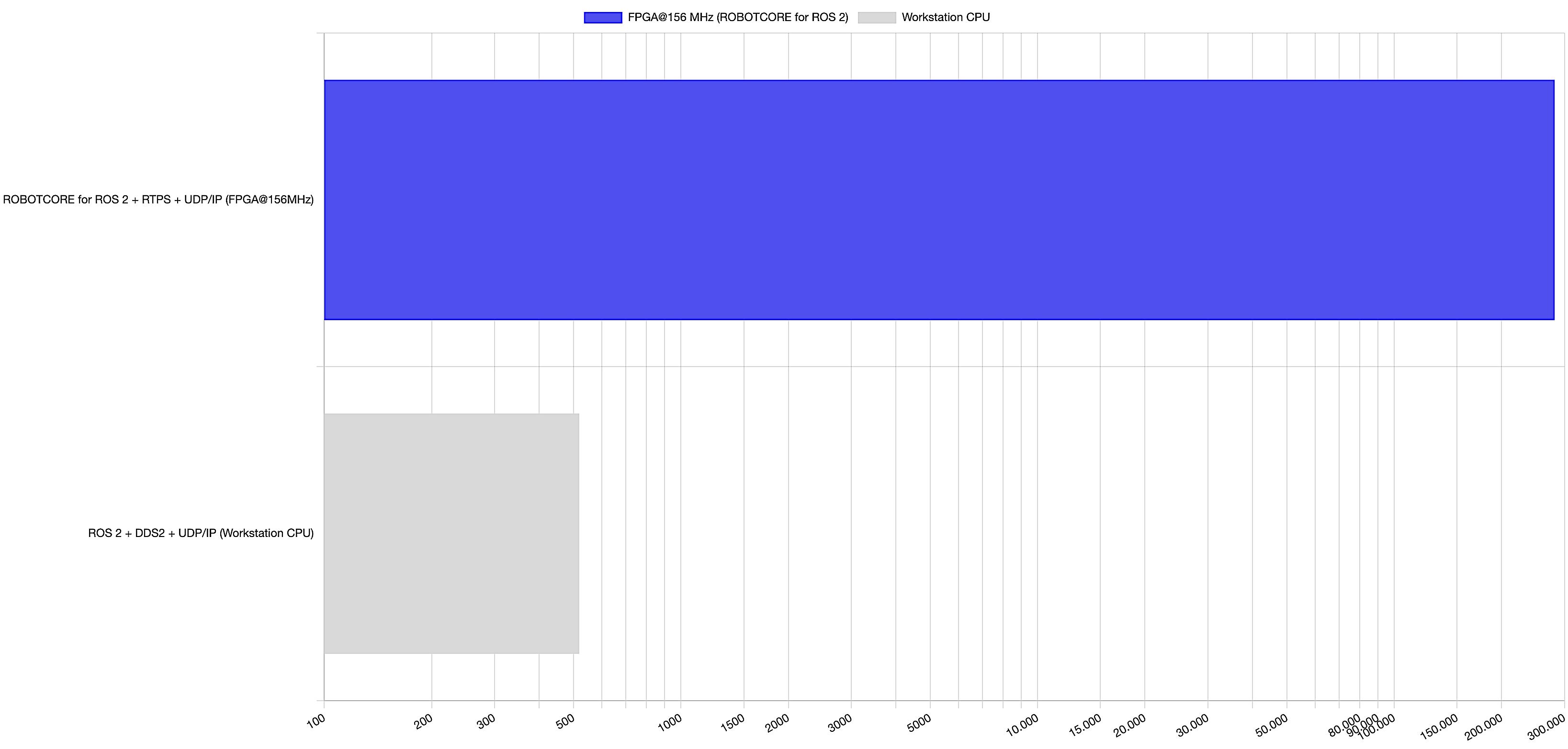}

    \setlength{\tabcolsep}{20pt}
    \renewcommand{\arraystretch}{1.5}
    \scalebox{0.6}{
    \begin{tabular}{  >{\centering\arraybackslash}m{11cm} |>{\centering\arraybackslash}m{7cm}|>{\centering\arraybackslash}m{5cm}} 
        
          & \color{black}\textbluebf{FPGA@156MHz \newline(}\texttt{ROBOTCORE for ROS 2 + RTPS + UDP/IP}\textbluebf{)} & \color{black}\textbf{Workstation CPU} \\
        \hline

        \rowcolor{black!5} Mean communication frequency-per-Watt (\texttt{energy efficiency}) & \textbluebf{281,690} (\texttt{543$\times$}) & {518} (\texttt{1$\times$}) \\
                
    \end{tabular}}
    \caption{Energy efficiency measured in terms of mean communication frequency-per-Watt. Logarithmic scale.}
    \label{fig:energy_efficient}
\end{figure}

Sending a ROS 2 message with \roschip takes about 1.775 microjoules (uJ). Our chip design stands out remarkably, outperforming contemporary workstation CPU and GPU solutions when it comes to power performance. \roschip exhibits an exceptional efficiency, being \texttt{500$\times$} more power-efficient compared to modern workstation CPUs and GPUs as depicted in figure \ref{fig:energy_efficient}. 

\noindent The comparison extends to the domain of communication frequency-per-Watt, a metric that encapsulates the balance between communication throughput and energy consumption. Our hardware configuration achieves an impressive mean communication frequency-per-Watt of \texttt{281,690}, signifying an efficiency that is \texttt{543$\times$} greater than that of conventional workstation CPUs, which register a mean of merely \texttt{518}. Such metrics not only highlight the hardware's adeptness at minimizing energy expenditure but also its capability to sustain high-frequency communications with minimal power input.

These findings, depicted in Figure \ref{fig:energy_efficient}, resonate with the broader objective of advancing robotic systems towards not only achieving real-time performance but doing so in an energy-conscious manner. The \texttt{ROBOTCORE} suite's ability to deliver such efficiency gains illustrates the tangible benefits of integrating specialized hardware for ROS 2 functionalities, paving the way for the development of more sustainable, energy-efficient robotic systems.

%
%

\section{Discussion and future work}
\label{sec:discussion}

\subsection{Achieving Brain-Like Speeds and Efficiency in Robotic Networking}



Sending a ROS 2 message with \roschip Design takes about 1.78 microjoules ($\mu$J) and about 2.5 microseconds ($\mu$s). For comparison, a single neuron of a human brain firing one action potential takes roughly 0.03 to 0.3 microjoules ($\mu$J) and lasts approximately 2 to 5 milliseconds (ms) \cite{pissadaki2013energy, levy2021communication}, while the energy required to Blink an Eye is roughly estimated to be around 1 millijoule (mJ) lasting about 300 to 400 milliseconds (ms) \cite{yang2024eye}. The remarkable efficiency and speed of the \roschip Design in exchanging data between robots not only brings robotic networking closer to biological processing speeds but also highlights the potential for robotics to emulate more closely the energy efficiency found in natural systems. This comparison not only underscores the advanced technological achievements of \roschip but also opens avenues for further exploration into making robotic systems that can operate within the energy budgets of biological counterparts. By reducing the energy consumption and latency in robotic communication, \roschip Design paves the way for the development of more responsive, autonomous systems that can perform complex tasks with a minimal energy footprint, mirroring the efficiency of biological organisms.


\subsection{Future work}

Our circuitry presented herein as \roschip  served within the ROBOTCORE® suite of IP operates with an energy budget comparable to the fundamental biological processes in the human brain and with much faster speeds. This is relevant for future research in both robotics and neuroscience, as \emph{"communication consumes 35 times more energy than computation in the human cortex"} \cite{levy2021communication}. In upcoming research, we'll focus on leveraging our specialized hardware chip design and benchmark it against industry use cases wherein these accelerated communication and computation capabilities can lead to faster robots in real scenarios.

Beyond industry, these advancement holds promise for enabling new applications in robotics where energy efficiency and rapid response times are critical, such as those requiring compute-demading levels of encryption, authentication and authorization. Further research could explore the integration of this  technology into broader networks of robots, exploring the inter-network level of interactions, creating collectives capable of complex, coordinated actions with efficiency and communication speeds that empower new applications.

\newpage


\bibliographystyle{IEEEtran}
\bibliography{bibliography}


\end{document}